\documentclass[letterpaper]{article} % DO NOT CHANGE THIS 
\usepackage{aaai24} %-> For camera-ready submission
\usepackage{times}  % DO NOT CHANGE THIS
\usepackage{helvet}  % DO NOT CHANGE THIS
\usepackage{courier}  % DO NOT CHANGE THIS
\usepackage[hyphens]{url}  % DO NOT CHANGE THIS
\usepackage{graphicx} % DO NOT CHANGE THIS
\usepackage{natbib}  % DO NOT CHANGE THIS AND DO NOT ADD ANY OPTIONS TO IT
\usepackage{caption} % DO NOT CHANGE THIS AND DO NOT ADD ANY OPTIONS TO IT
\usepackage{algorithm}
\usepackage{algorithmic}

\usepackage{newfloat}
\usepackage{listings}
\DeclareCaptionStyle{ruled}{labelfont=normalfont,labelsep=colon,strut=off} % DO NOT CHANGE THIS
\floatstyle{ruled}
\newfloat{listing}{tb}{lst}{}
\floatname{listing}{Listing}
\usepackage{xcolor}
\usepackage{booktabs}
\title{ESG Accountability Made Easy: DocQA at Your Service}

\author{
    Lokesh Mishra \textsuperscript{\rm 1},
    Cesar Berrospi \textsuperscript{\rm 1},
    Kasper Dinkla \textsuperscript{\rm 1},
    Diego Antognini \textsuperscript{\rm 1},
    Francesco Fusco \textsuperscript{\rm 1},
    Benedikt Bothur \textsuperscript{\rm 2},
    Maksym Lysak \textsuperscript{\rm 1},
    Nikolaos Livathinos \textsuperscript{\rm 1},
    Ahmed Nassar \textsuperscript{\rm 1},
    Panagiotis Vagenas \textsuperscript{\rm 1},
    Lucas Morin \textsuperscript{\rm 1,3},
    Christoph Auer \textsuperscript{\rm 1},
    Michele Dolfi \textsuperscript{\rm 1},
    Peter Staar \textsuperscript{\rm 1}
}
\affiliations{
    \textsuperscript{\rm 1}IBM Research, R\"uschlikon, Switzerland\\
    \{mis, ceb, dkl, ffu, mly, nli, ahn, pva, lum, cau, dol, taa\}@zurich.ibm.com\\
    \textsuperscript{\rm 2}IBM Technology, Zürich, Switzerland\\
     \{Benedikt.Bothur, Diego.Antognini\}@ibm.com \\
     \textsuperscript{\rm 3}ETH Zürich, Zürich, Switzerland\\

}

\begin{document}

\maketitle

\begin{abstract}
We present Deep Search DocQA. This application enables information extraction from documents via a question-answering conversational assistant. The system integrates several technologies from different AI disciplines consisting of document conversion to machine-readable format (via computer vision), finding relevant data (via natural language processing), and formulating an eloquent response (via large language models). Users can explore over 10,000 Environmental, Social, and Governance (ESG) disclosure reports from over 2000 corporations.
The Deep Search platform can be accessed at: \url{https://ds4sd.github.io}.
\end{abstract}

\section{Introduction}

The global impact of climate change has galvanized organizations to announce key information about their environmental footprint (carbon emissions, energy usage, waste emission and management, etc.). Integrating sustainability information into the company reporting cycle is one of the targets of the UN 2030 Agenda for Sustainable Development
% ~\cite{sdgs2030}
and institutions like Principles for Responsible Investing (a UN-supported network of investors) encourage investors to incorporate this information into their investment decisions
% ~\cite{pri}
. Companies are thus increasingly disclosing environmental, social, and governance (ESG) data in their ESG reports, typically as PDF files.

% There have been several massive efforts from different international organizations to standardize these reports, such as: recommendations from Task Force on Climate-related Financial Disclosures (TCFD), European Sustainability Reporting Standards (ESRS), Sustainability Accounting Standard Board (SASB), Global Reporting Initiative (GRI), Carbon Disclosure Project (CDP), etc. 
Unlike financial data, regulators such as the U.S. SEC do not require public companies to file ESG data with specific forms. There have been massive efforts from several organizations to standardize these reports. However, major challenges continue to persist, including complex regulations, rapidly evolving reporting frameworks, verifying ESG compliance, among others. These matters become more complicated when we realize that most of the ESG reporting is done in non machine-readable formats. Unlocking this vast amount of data in an easily consumable manner would greatly help researchers, policy-makers, lawyers, and corporations by extracting information and gaining insights.

To this end, we have developed Deep Search \mbox{DocQA}. The application offers users to perform document \textit{question answering} (QA), i.e., users can extract information from any ESG report in our library via our QA conversational assistant. Our assistant generates answers and also presents the information (paragraph or table), in the ESG report, from which it has generated the response. 
% The ESG collection in our library consists of over 10,000 ESG reports, from about 2000 corporations, between the years 1992 and 2023.

\begin{table}[!t]
\scriptsize
\begin{tabular}{p{0.9cm}|p{3.3cm}|p{3cm}}
\toprule
\textbf{Report} & \textbf{Question} & \textbf{Answer} \\
\midrule
IBM 2022 &How many hours were spent on employee learning in 2021? & 22.5 million hours\\
\midrule
IBM 2022 & What was the rate of fatalities in 2021? & The rate of fatalities in 2021 was 0.0016.\\
\midrule
IBM 2022 & How many full audits were conducted in 2022 in India? & 2\\
\midrule
Starbucks 2022 & What is the percentage of women in the Board of Directors? & 25\%\\
\midrule
Starbucks 2022 & What was the total energy consumption in 2021? & According to the table, the total energy consumption in 2021 was 2,491,543 MWh.\\
\midrule
Starbucks 2022 & How much packaging material was made from renewable materials? & According to the given data, 31\% of packaging materials were made from recycled or renewable materials in FY22.\\
\bottomrule
\end{tabular}
\caption{Example question answers from the ESG reports of IBM and Starbucks using Deep Search DocQA system.}
\label{tab:qa-examples}
\end{table}

\begin{figure*}[t]
\centering
\includegraphics[width=0.9\textwidth]{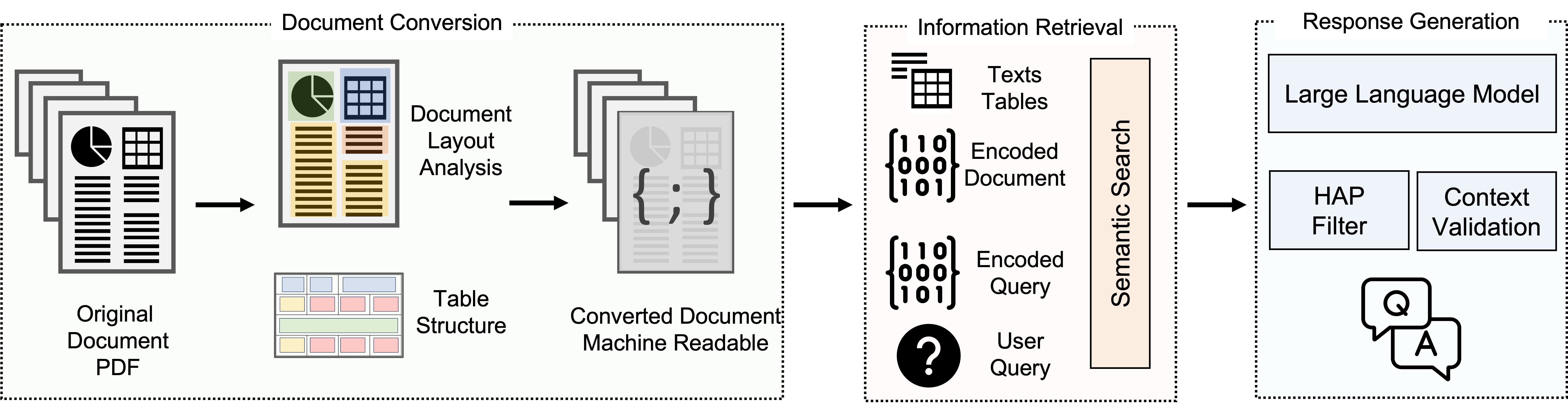} % Reduce the figure size so that it is slightly narrower than the column.
\caption{System architecture: Simplified sketch of document question-answering pipeline.}
\label{fig:pipeline}
\end{figure*}

\section{Related Work}
The DocQA integrates multiple AI technologies, namely: 
% Our discussion of related work is split along the following axes: document conversion, application of NLP to ESG reports, and development of large language models.

 % for converting data in documents to a machine-readable format, understanding table structure, semantically identifying data relevant to a question, generating an answer from a relevant context etc.
\textbf{Document Conversion: } Converting unstructured documents, such as PDF files, into a machine-readable format is a challenging task in AI.
% This task is particularly complicated because converting a document to a machine-readable format is difficult due to the large inherent svariety in documents and their structures. 
Early strategies for document conversion were based on geometric layout analysis~\cite{cattoni1998_geometric, breuel2002}. Thanks to the availability of large annotated datasets (PubLayNet~\cite{zhong2019_publaynet}, DocBank~\cite{li2020_docbank}, DocLayNet~\cite{pfitzmann2022_doclaynet, Auer_2023}, deep learning-based methods are routinely used. Modern approaches for recovering the structure of a document can be broadly divided into two categories: \textit{image-based} or \textit{PDF representation-based}. Image-based methods usually employ Transformer or CNN architectures on the images of pages~\cite{Zhang2023_WeLayoutWL,Li2022DiTSP,Huang2022LayoutLMv3PF}. On the other hand, deep learning-based language processing methods are applied on the native PDF content (generated by a single PDF printing command) \cite{Auer_2022, livathinos2021robust, Staar2018}. 

\textbf{Application of NLP to ESG: } ESG reports contain large amount of useful data in textual and tabular format. There have been some attempts to use NLP on this data. \citet{luccioni_analyzing_2020} developed ClimateQA, a model trained to classify whether a sentence from an ESG report answers regulatory questions. 
% CLIMABENCH proposes benchmark datasets for the tasks of QA ranking and classification \cite{laud_climabench_2023}. 
In addition, there are several works which aim to mine information from ESG reports for financial predictions \cite{guo2020esg2risk, Goel2020MiningCS}. Nevertheless, to the best of our knowledge, no QA system which can extract data directly from an ESG report PDF has been reported in the literature.

\textbf{LLM \& RAG: } 
% Self-supervised language modelling engendered foundation large language models \cite{yang2023_harnessing, bommasani2022_foundationmodel, Radford2019_gpt2, raffel2020_t5, devlin2019_bert}. 
Due to the increasing scale of training data and model size, large language models (LLMs) demonstrate surprising emergent properties~\cite{wei2022_emergent}. For example, the behaviour of the GPT-3 model, with 175 billion parameters, could be modified with in-context learning~\cite{brown2020_gpt3,bommasani2022_foundationmodel, raffel2020_t5}. Such LLMs are adaptable to a variety of downstream tasks via prompting and can be fine-tuned to better perform in a specific ESG domain~\cite{wkbl2022climatebert}. The Retrieval Augmented Generation (RAG) approach aims at improving the performance of these models on knowledge intensive tasks \cite{lewis2021retrievalaugmented}. In this approach, the capabilities of natural language generation are combined with a knowledge index, from which relevant documents are retrieved.

\section{System Architecture}

In this section, we describe the AI technologies which are integrated into our document question-answering application. The architecture is described in Fig.~\ref{fig:pipeline}. The pipeline works end-to-end from PDF documents to question-answering using LLMs. It consists of three components described below. 

\textbf{Document Conversion:} The document conversion system is designed in an asynchronous task-based queue-worker architecture. The user-facing API accepts documents in PDF format (both programmatically created and scanned). The client receives a task identifier, while an orchestrator enqueues several ML tasks to ephemeral workers. After splitting the document into pages, we: 1) depending on the nature of the PDF, we employ either PDF parsing or OCR, 2) analyze layout and segment it~\cite{Auer_2022, livathinos2021robust, Staar2018} and 3) extract table structures~\cite{lysak_optimized_2023, nassar_tableformer_2022}. Finally, the data from multiple pages is assembled together, preserving the reading order, in a machine-readable format.

\textbf{Information Retrieval:} Using an encoder model, vector embeddings for the data in a document are computed and stored in a vector database.
% \footnote{For our database we use Redis \url{https://redis.io/}}. 
For text this is relatively straightforward, for tables the triplet of (cell content, column header, row header) is expressed as a sentence which gets encoded. The sentence expression is: \texttt{string(column header) + string(row header) = string(cell content)}\footnote{Here, string() returns the string representation of an object.}. We perform a k-nearest neighbour search to identify the top-k relevant passages for a user query. For sentence encoding, we use several encoding models from the Sentence Transformer library~\cite{reimers2019sentencebert}. 

\textbf{Response Generation:} We employ a suite of LLMs like LLAMA 2~\cite{touvron2023llama}, Flan-UL2~\cite{tay2023ul2}, or T5~\cite{raffel2020_t5} for generating a response to the user query. The user query and relevant context (identified by the previous model) are packaged together in a prompt for the LLM. The response of the model is checked against hate speech, abuse, and profanity. Finally the response is grounded in the context and inspected for hallucinations. If all tests are passed, the response is presented to the user via a virtual assistant. Table~\ref{tab:qa-examples} shows some examples of questions and the generated answers by the system.

\section{Conclusions}

In this paper, we presented our DocQA application targeting ESG reports. The DocQA system can be useful for anyone, from policy-makers to students, trying to find information from a large document. Our future work is focused on enabling querying on multiple documents at once to extract aggregated insights for questions like ``How have the Scope 1 emissions evolved over the last decade?". In addition, we will expand this service to other types of documents like scientific papers, financial reports, and patents.

\bibliography{main}
\end{document}